\documentclass[11pt,a4paper]{article}
\usepackage[utf8]{inputenc}
\usepackage{babel}

\usepackage{lineno,hyperref}

\usepackage{smartdiagram}
\usesmartdiagramlibrary{additions}
\usepackage{amsmath}
\usepackage{multirow}
\usepackage{algorithm} 
\usepackage{float}
\usepackage[caption = false]{subfig}
\usepackage{graphicx}
\usepackage[noend]{algpseudocode}
\usepackage{authblk}
\usepackage{orcidlink}

\title{Active Learning and Transfer Learning for Anomaly Detection in Time-Series Data}

\author[1]{John D. Kelleher \orcidlink{0000-0001-6462-3248}} 
\author[1]{Matthew Nicholson} 
\author[1]{\authorcr Rahul Agrahari} 
\author[2]{Clare Conran} 

\affil[1]{ADAPT Research Centre, School of Computer Science and Statistics, Trinity College Dublin, Ireland}
\affil[2]{ReliaQuest, Cambridge, UK}
\affil[ ]{\texttt{john.kelleher@tcd.id}}

\date{}

\begin{document}
\maketitle

\begin{abstract}
This paper examines the effectiveness of combining active learning and transfer learning for anomaly detection in cross-domain time-series data. Our results indicate that there is an interaction between clustering and active learning and in general the best performance is achieved using a single cluster (in other words when clustering is not applied). Also, we find that adding new samples to the training set using active learning does improve model performance but that in general, the rate of improvement is slower than the results reported in the literature suggest. We attribute this difference to an improved experimental design where distinct data samples are used for the sampling and testing pools. Finally, we assess the ceiling performance of transfer learning in combination with active learning across several datasets and find that performance does initially improve but eventually begins to tail off as more target points are selected for inclusion in training. This tail-off in performance may indicate that the active learning process is doing a good job of sequencing data points for selection, pushing the less useful points towards the end of the selection process and that this tail-off occurs when these less useful points are eventually added. Taken together our results indicate that active learning is effective but that the improvement in model performance follows a linear flat function concerning the number of points selected and labelled.
\end{abstract}

\section{Introduction}

Recently, there has been a huge increase in the usage of cloud services across the world. Nearly all the software applications are deployed on clouds. These cloud systems are used by millions of people and so these services must have very high service availability. Consequently, the early detection and resolution of anomalies in these systems are of critical importance to the successful deployment of applications. However, modern cloud systems are a combination of multiple services and each service produces enormous volumes of monitoring data which makes it extremely difficult to analyse this monitoring data for anomalies. There is a growing body of research on identifying the faults in the cloud system, often this work is framed in terms of anomaly detection and uses supervised machine learning techniques. 

Supervised machine learning, however, requires access to labelled training data. The creation of labelled datasets is a labour-intensive and expensive task, and creating new training datasets for each new system and associated stream of monitoring data is not a scalable strategy. A potentially feasible solution to the problem of acquiring labelled data for model training is to leverage transfer learning and active learning. However, relatively little work has examined how active learning and transfer learning work in combination. In this paper, we assess how well active learning and transfer learning work in combination.

In previous work, we examined the effectiveness of transfer learning for anomaly detection in cloud services \cite{nicholsonetal:2022}. In this paper we focus on active learning and assess three questions:

\begin{enumerate}
	\item Do the optimal parameters for a transfer learning system remain stable or change when transfer learning is combined with active learning? We examine this question in terms of clustering hyper-parameters. 
	\item At what rate can we expect model performance to improve as active learning progresses? 
	\item What happens as the number of samples selected using active learning grows? 
\end{enumerate}

The paper is structured as follows: we first provide an introduction to active learning, we then describe the datasets used in our experiments and our experimental setup (data splits and evaluation metrics), we then report the results from three different experiments, and the paper finishes with a summary and conclusions section. 

\section{Active Learning}

Active learning is based on the belief that comparable model performance can be achieved using a small, curated dataset as compared to a large dataset. Building on this belief, the goal of active learning is to reduce the cost of data labelling by attempting to select the most useful data points to label to achieve high model accuracy. The basic idea is to iteratively train models on a task using small amounts of labelled data, use the model performance on the data to inform the selection of new data points for labelling, and then label the selected data points and retrain the model using the extended training dataset. In essence, the model is used to inform the selection of the data that is used to retrain the model.

Broadly speaking there are three main problem scenarios in which active learning can be applied these are: 
\begin{enumerate}
	\item Stream Based Selective Sampling: in this scenario, the unlabelled data points are examined one at a time with the active learning process evaluating the informativeness of each data point individually. The learner decides for itself whether to assign a label or query the teacher for each data point.
	\item Pool-Based Sampling:  in this scenario, instances are drawn from the entire data pool and assigned an informativeness score. Often these informativeness scores are a function of the uncertainty of the system on classifying a point, the intuition being the higher the system uncertainty for a point the more informative the inclusion of that point in the training data. The system then selects the most informative instances and queries the teacher for the labels. Note that by considering the data as being selected from a pool, rather than individually, the concept of data diversity can be introduced into the data selection process. 

	\item Membership Query Synthesis: This is where the learner generates an instance from an underlying natural distribution. This is particularly useful if the dataset is small. See \cite{wangetal2015} for an example of membership query synthesis work.
\end{enumerate}
Of these three active learning scenarios, the one most relevant to this work is pool-based active learning, which is illustrated in Figure~\ref{fig:activelearning} below.

\begin{figure}
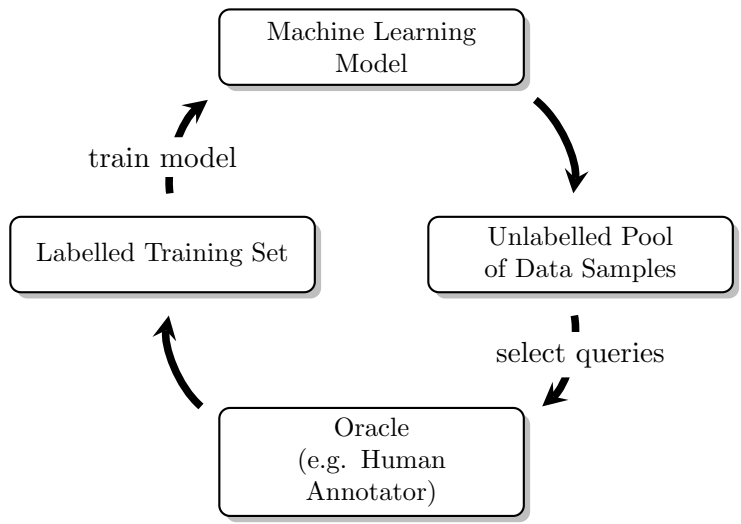

\centerline{
\smartdiagramset{
	module minimum width=4cm, 
	text width=3.75cm, 
	uniform color list=white for 4 items, 
	border color=black, 
	uniform arrow color=true, 
	arrow color=black,
	additions={
		additional item width=2.5cm,
		additional item height=0.2cm,
		additional item text width=2.5cm,
		additional item offset=0.5cm,
		additional item border color=none,
		additional item fill opacity=1,
		additional item fill color=white,
	}
}
\smartdiagramadd[circular diagram:clockwise]{
Machine Learning\\Model, Unlabelled Pool\\of Data Samples, Oracle\\(e.g. Human Annotator), Labelled Training Set
}{
below of module2/{select queries},above of module4/{train model}
}
}
\label{fig:activelearning}
\caption{The pool-based active learning cycle, inspired by Figure 1 in \cite{settles2009active}}
\end{figure}

Active learning systems use an \emph{acquisition function} to select data points from a set of unlabelled data. The design goal of these acquisition functions is to select points whose inclusion in the training data will maximise model performance.  In pool-based active learning scenarios acquisition function can be designed to balance two considerations: 
\begin{enumerate}
	\item maximising information gained from selected points, 
	\item minimising redundant information from similar data points. 
\end{enumerate}
Selecting data points in batches is common to reduce computation and time required for training. 

The goal of the research reported in this paper is to assess the likely efficiency of active learning in terms of improvement in model performance relative to data labelling cost for anomaly detection in cloud systems time-series datasets. A relevant recent example of this type of research is the framework for cross-domain cloud series time-series anomaly detection proposed in \cite{zhang2019cross} which combines transfer learning with active learning. Because the active learning approach proposed in \cite{zhang2019cross} is specifically designed for the same context that we are interested in assessing active learning within, in the experiments we report below we use the acquisition function proposed by \cite{zhang2019cross} as representative of the likely effectiveness of active learning in this context of use.   

The acquisition function of \cite{zhang2019cross} integrates two components:
\begin{enumerate}
	\item a measure of a model's uncertainty on a data point as an indicator of the informativeness of the data point. This component is implemented as a rank order over the data points in the unlabelled pool of data samples, with items with high uncertainty at the top of the list. The first step in creating this rank order of data points is to calculate for each data point the predicted probability of that item being an anomalous item $P(anom)$, as judged by an anomaly detection machine learning model. The probability of a data point being a normal data point can then be calculated as $P(norm)=1-P(anom)$. The certainty of the model regarding a data point is then calculated as the absolute difference between $P(norm)$ and $P(anom)$: $|P(norm)-P(anom)|$. Finally, the data points in the unlabelled pool are ranked in increasing order of certainty.  This means that items that have a small absolute difference between $P(norm)$ and $P(anom)$ are considered to have high uncertainty and are considered most informative to labelling and retraining the model, and so are ranked at the top of the selection list. 
	\item and a measure of context diversity to minimise redundancy in the data selected for annotation (i.e., removing points that occur within a time window of another selected point within a time series, assuming points that occur in close vicinity within a time series will be similar and so contribute redundant information to the learning process). In  \cite{zhang2019cross},  the context of sample $x_t$ is controlled by a parameter $\alpha$, which defines a range from $x_t - \alpha$ to $x_t + \alpha$ in a time series. Consequently, after a point with high uncertainty has been selected from the rank-ordered listed for labelling if a point later in the list is being considered for selection is within the range of context diversity (i.e., is adjacent within the time-series) this later point is not be considered for selection. In other words, the context diversity measure acts as a filter on the selection of data points for labelling.
\end{enumerate}

The framework proposed in \cite{zhang2019cross} combines the active learning process described above with transfer learning. The process begins by training a base model using a labelled dataset from a similar domain to the domain into which the model will be deployed. This initial training dataset and base model provide the basis for the active learning process. The process also requires a sample of unlabelled data from the target domain into which the model will be deployed. Given these inputs the active learning process runs for a predetermined number of iterations and in each iteration completes the following sequence of steps:
\begin{enumerate}
	\item The base model is run on the pool of unlabelled target data points and its scores are used to rank the data points in terms of uncertainty 
	\item A pre-set number of data points from the unlabelled target domain are selected such that: (a) they are the samples that have the highest uncertainty, and (b) none of the selected data points are in the context window of other selected data points
	\item The selected data points are labelled and added to the training dataset.
	\item The base model is retrained and this new model is used in the next iteration of the algorithm
\end{enumerate}

The algorithm finishes once the predetermined number of iterations has been completed. The output of the process is the final version of the model after it has been retrained at the end of the last iteration. 

Although in our experiments we use the active learning algorithm from \cite{zhang2019cross} as representative of active learning approaches for anomaly detection in cloud-series time-series an important difference between how we assess the effectiveness of active learning and how the experiments reported in \cite{zhang2019cross} were designed is that whereas in our experiments we use separate pools of data points from the target domain for active learning sampling versus model evaluation (i.e., one sample from the target domain is used to select points for labelling and model retraining and another sample is used as a test set). This differs from the experiments in \cite{zhang2019cross} where the same sample of data was used for both of these tasks. Consequently, in the experiments in \cite{zhang2019cross}, each iteration of active learning results in the test set becoming smaller (by the fact that the points selected for labelling and inclusion in the training data are removed) and easier (because, in general, the active learning process will remove the data points from the test set that the model has the highest uncertainty for). By contrast, in our experiments, we maintain the same test set throughout and apply active learning to a different sample of target data. This means that it is to be expected that the rate of performance improvement in our experiments with respect to the number of points labelled will be slower as compared to those reported in \cite{zhang2019cross}, but also that the metrics we report in terms of the model's ability to generalise to new unseen data (i.e., data not used directly in the training process) are, in our view, more valid.

\section{Datasets, Experimental Setup and Evaluation Metrics}

We use six datasets in our experiments: NAB (AWS and Twitter) \cite{lavin2015evaluating}, Yahoo (Real and Artificial) \cite{laptev2015generic}, IOPS KPI\footnote{Available from: \url{https://github.com/NetManAIOps/KPI-Anomaly-Detection}}, and Huawei\footnote{From the 2020 Huawei anomaly detection  competition: \url{https://huawei-euchallenge.bemyapp.com/ireland}}. Each dataset contains multiple files and each file contains one time series. Table~\ref{tab:Dataset-Description} provides summary statistics for each dataset. Furthermore, following the results reported in \cite{agraharietal:2022} we use the catch24 feature set (and extension of the catch22 dataset from \cite{lubba2019}) in combination with random forest models for all our experiments.  

Note that the focus of this paper is on understanding the interactions between active learning and transfer learning and on assessing the general effectiveness of active learning rather than on fine-tuning the performance of models. Consequently, we did not expend a large amount of time on fitting model hyper-parameters. Instead, we selected a single set of hyper-parameters for the random forest model (these were the default hyper-parameters set for \emph{sklearn} \cite{scikit-learn}, the machine learning library we used to implement our experiments) and kept these consistent across the datasets:  $max\_features=\sqrt{n\_features}$, no max depth, two samples as a minimum number of samples required to split an internal node, one as the minimum number of samples in newly created leaves, with~bootstrapping, using out-of-bag samples to estimate the generalization error.

\begin{table}
 \begin{tabular}{ lcccc  }
 \hline
Dataset& \# of Points &\% of Anomalies& \# of Time Series&Mean Length\\
 \hline
 Yahoo Real & ~91k & 1.76\% & 64 & 1415\\ 
 Yahoo Artificial & 140K & 1.76\% & 100 & 1415\\ 
 IOPS & ~3M & 2.26\% & 29 & 105985\\ 
 AWS & ~67K & 4.57\% & 17 & 67740\\ 
 Twitter & ~142K & 0.15\% & 10 & 142765\\ 
 Huawei & ~54K & 4.19\% & 6 & 9056\\ 
\hline
\end{tabular}  
\caption{Summary statistics for the datasets used in the experiments}
\label{tab:Dataset-Description}
\end{table}

Our experiments assess the performance of active learning in the context of cross-domain anomaly detection. Consequently, the basic structure of our experimental design involves transfer learning. To carry out a transfer learning experiment we need to define a target and source dataset. To do this we treat each of the 6 datasets as a target dataset in turn and a new \emph{multi-source dataset} is created by merging the other five datasets. This process of data handling resulted in the following transfer combinations (source $\rightarrow$ target): 
\begin{enumerate}
	\item Non-Aws $\rightarrow$ Aws
	\item Non-Huawei $\rightarrow$ Huawei
	\item Non-IOPS $\rightarrow$ IOPS
	\item Non-Twitter $\rightarrow$ Twitter
	\item $\text{Non-Yahoo}_{\text{Artificial}} \rightarrow \text{Yahoo}_{\text{Artificial}}$
	\item $\text{Non-Yahoo}_{\text{Real}} \rightarrow \text{Yahoo}_{\text{Real}}$
\end{enumerate}

The Non-X part consists of all the datasets except for the mentioned name (i.e., all datasets except for X) and it is the source part of the experiment. The counterpart is the target dataset which consists of the mentioned dataset. The IOPS dataset is significantly larger than the rest of the datasets and so a 5\% stratified sample, ensuring the same proportion of anomalies is present, was taken when it was used as a source dataset. The full IOPS dataset is used when it is used as the target dataset. 

Experiments reported in \cite{nicholsonetal:2022} found that in the context of transfer learning for anomaly detection it can be beneficial to apply clustering to the target data to identify sub-domains with the target domain and then to train a separate anomaly detection model for each of these sub-domains. Inference then involves assigning each test data point to a target sub-domain and processing the test point with the corresponding anomaly detection model. We adopt the same transfer learning data processing pipeline in our experiments reported in this paper. Accordingly, for each transfer combination of datasets listed above, once the source and target datasets are created we fit a K-means++ model on the target dataset for various numbers of k where k is the number of clusters (for each experiment below we will report the specific values of k that were used). After fitting the kmeans model to the target, the model is applied to the corresponding source dataset. The outcome of this process is a set of clusters where each cluster contains data points from the target and the corresponding source domain. Then for each cluster, a 5-fold cross-validation process is applied to get the train and test split for 5 folds. The folds are sampled such that the percentage of anomalies is constant across folds, in other words, the percentage of target anomalies is constant across each fold and the percentage of source anomalies is constant across each fold. This process creates for each fold: a source training set, a source test set, a target training set and a target test set. It is the fact that we create a separate training and test set for the target domain that enables us---unlike \cite{zhang2019cross}---to maintain the same test set across the iterations of the active learning while we apply the active learning process (in terms of selecting data points, labelling and integration into the model training set) solely to the training split of the target domain.

For each fold, the source data and the training part of target data are fitted on the kmeans models to distribute the points across the multiple clusters. For each cluster, the source data is transformed to be more similar to the target training data point in the cluster using CORAL \cite{sun2016return} (a domain adaptation technique designed to reduce the difference between source and target) and is trained on ML model for the anomaly detection task. Once this base model has been trained using the adapted source domain we are ready to begin the experiments on active learning. 

\begin{figure}
\includegraphics[width=\textwidth]{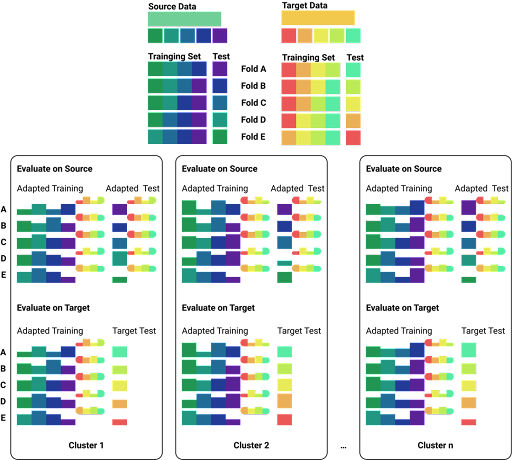}
\caption{Illustration of the definition of the source and target training and test sets for the 5-fold cross-validation. Note each source set is only adapted once using CORAL, e.g. source training set A is only adapted using target training set A, not B, C, D, or E. Best viewed in colour.}
\label{fig:experimentaldesign}
\end{figure}

In all of our experiments, the base model is a random forest model. Also, to reiterate, the target points selected via active learning for inclusion in the training data are always sampled from the target domain training set (not the test set). The active learning process used is the approach described above and proposed by \cite{zhang2019cross}. The first points from the available set are ranked according to uncertainty. Then several points are selected, excluding points that are within a $\pm10$ window of already selected points. This process is repeated for 5 rounds. After each round models and the uncertainties used in the next round are based on the predictions of the retrained models. Figure~\ref{fig:experimentaldesign} illustrates the experimental design we use to divide the data into 5 folds and define source and target training and test sets. One point of note is that we solely focus on the effectiveness of active learning in the context of transfer learning from multiple combined source domains, and so in our experiments, our analysis reports on the performance and change in performance on the test set target dataset as the active learning process proceeds. 

In our experiments model accuracy is evaluated using F1 (Equation \ref{eq:f1}), Precision (Equation \ref{eq:p}), and Recall (Equation \ref{eq:r}):

\begin{equation}
F1=\frac{2 \times P \times R}{P + R}
\label{eq:f1}
\end{equation}

\begin{equation}
P = \frac{TP}{TP + FP}
\label{eq:p}
\end{equation}

\begin{equation}
R= \frac{TP}{TP + FN}
\label{eq:r}
\end{equation}

where P and R are the precision and recall respectively. TP is the true positive (Anomalous), TN is the true negative(Non-Anomalous) and FP, and FN are false positive and false negative. Our main focus is on the F1 score as it keeps the balance between Precision and Recall. For each experiment, we report the mean value for each of these metrics across the 5 folds.  

\section{Experiment 1: Understanding the interaction between Clustering and Active Learning}
\label{sec:exp1}

As we noted above the experiments reported in \cite{nicholsonetal:2022} found that the best transfer learning-based model performance for anomaly detection in time-series data occurred when clustering on the target data. In this experiment, we investigated the interaction between clustering for transfer learning and active learning. An interaction of this type is likely present because as the number of clusters increases so too does the number of models trained, and this will result in the new training data points selected generated via active learning being distributed across the datasets for different models which are likely to dilute the impact of each round of active learning on model performance. Furthermore, as the number of clusters increases, there is also a tendency for an imbalance in cluster sizes to develop (e.g., clustering can return one or two large clusters with several smaller clusters). In the context of anomaly detection where the distribution of labels is imbalanced this fragmentation of the data can exacerbate the label imbalance (this is likely particularly true for small clusters which may have no positive--anomalous--examples, or have a disproportionately large number of positive samples). This in turn can result in the models trained on these clusters being overly confident in their predictions thereby undermining the uncertainty measure used by the active learning acquisition function. In summary, clustering can affect active learning both in terms of which points are selected by the active learning process and also in terms of the rate of improvement of overall system performance concerning the number of data points selected and labelled. 

Consequently, in this experiment, we investigate how the number of clusters affects model performance as data points are added to the training data using active learning. For each of our 6 transfer learning dataset combinations (e.g. Non-AWS $\rightarrow$ AWS, Non-Huawei $\rightarrow$ Huawei, and so on) we ran this experiment across 5 folds of cross-validation. Within each fold, we fix the number of iterations of active learning to be 5. However, we vary the number of points added across these 5 iterations of active learning across the following values: 0, 10, 20, 40, 80, 160, 320, 640, and 1280. What this means for example is that if we are adding 20 points across the 5 interactions then 20/5=4 data points are selected from the target training split and added to the model training dataset in each iteration. Algorithm~\ref{alg:exp1} lists a pseudocode description of this experimental design. Note that in this pseudocode N denotes the total number of points that are added across the 5 iterations of active learning that occur within each fold.

\begin{algorithm}
	\begin{algorithmic}[0]
		\State Select a dataset as the target, merge other datasets to form the source dataset
		\For{N in \{0, 10, 20, 40, 80, 160, 320, 640, 1280\}}
	    		\For{k in {1,$\dots$,10}}
				\State Fit k-means++ model on the target dataset to identify k clusters;
            			\State Use stratified 5-fold sampling to form training and test sets for both 
				\State source and target datasets; 
            			\For{each fold in 5-fold cross validation}
                				\State Apply k-clustering model to source and target training/test data;
                				\For{each of the k clusters}
                    				\State Fit CORAL using source and target training data in the cluster;
                    				\State Apply CORAL transformation to source training and test data 
						\State in the cluster to form adapted source training and test sets;
                    				\State Fit anomaly detection model on adapted source training set;
					\EndFor
					\For{5 rounds of active learning}
						\For{each of the k clusters}	
                    					\State Evaluate the points in the target training pool based on
							\State uncertainty of the cluster's anomaly detection model;
							\State Working from highest uncertainty down sequentially 							\State select the N/5 points in the target training data that have 							\State the highest uncertainty but which are not in the context 							\State window $\pm \alpha$ of any of the previously selected points;
							\State Label the newly selected points, add them to the training
							\State data and retrain the detection model for the cluster;
							\State Evaluate the retrained model on the points in the target							\State test set that are in the cluster;
                					\EndFor
            				\EndFor
        				\EndFor
			\EndFor
		\EndFor
    	\State For each dataset, number of clusters (k), and total quantity of points added by active learning (N) calculate an average precision, recall and F1 on the target test sets across the 5 folds. 
	\end{algorithmic} 
\caption{Experiment 1 Methodology}
\label{alg:exp1}
\end{algorithm}

\begin{figure}
\subfloat[AWS]{\includegraphics[width = 0.5\textwidth]{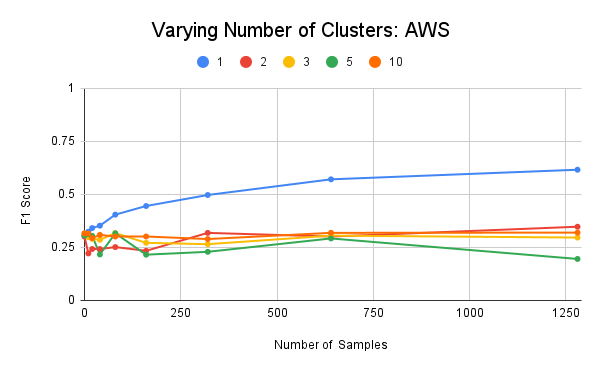}} 
\subfloat[Huawei]{\includegraphics[width = 0.5\textwidth]{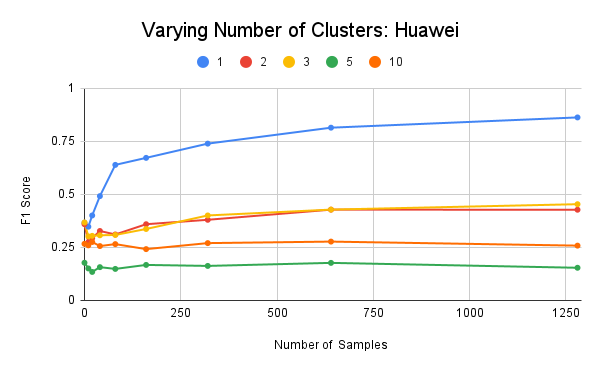}}\\
\subfloat[IOPS]{\includegraphics[width = 0.5\textwidth]{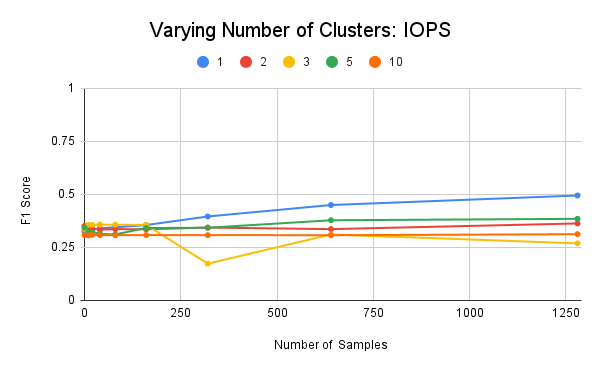}}
\subfloat[Twitter]{\includegraphics[width = 0.5\textwidth]{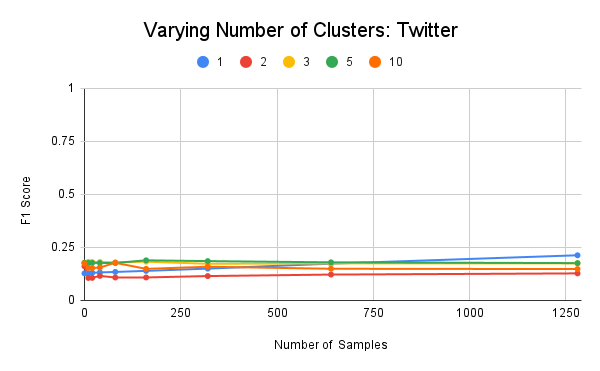}}\\ 
\subfloat[Yahoo Artificial]{\includegraphics[width = 0.5\textwidth]{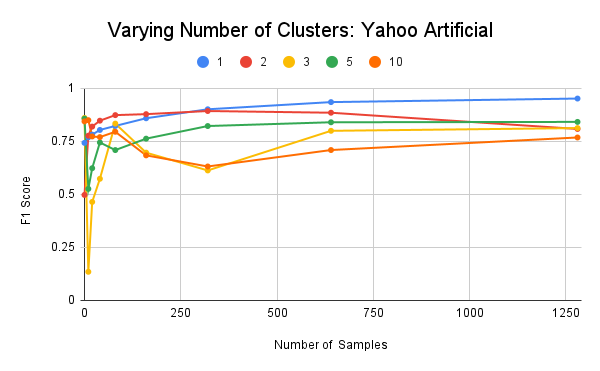}}
\subfloat[Yahoo Real]{\includegraphics[width = 0.5\textwidth]{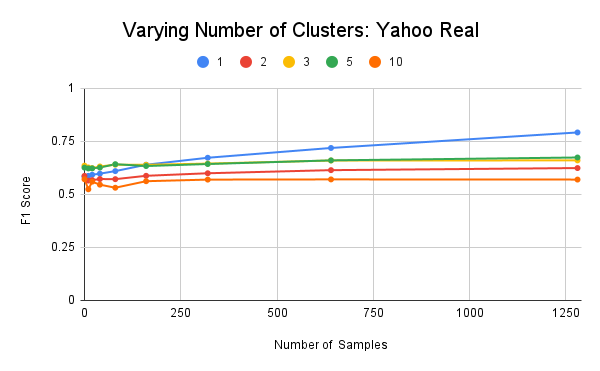}}\\ 
	\caption{For each dataset a graph illustrating the progression in F1 scores as more points are added via active learning when different numbers of clusters are used. Best seen in colour.}
	\label{fig:exp1-results}
\end{figure}

Figure~\ref{fig:exp1-results} shows for each dataset a graph illustrating the progression in F1 scores as more points are added via active learning when different numbers of clusters are used. There is a clear trend that active learning performs the best when there is a single cluster/model (in most cases the plot of a single cluster--blue line--is at the top of the graph). Note that before active learning the optimal number of clusters can vary (which is in line with the results of \cite{nicholsonetal:2022}), however for all datasets by the time 1,280 points have been selected via active learning the best performant model is the one a single cluster. This shift in performance to a single cluster is very evident in Table~\ref{tab:exp1f1bycluster} which lists for each dataset the before and after active learning performance by cluster. These results indicate that, as hypothesized, there is a strong interaction between active learning and the clustering processes used during transfer learning and that when active learning is applied best performance tends to occur when clustering is removed from the pipeline. 

\begin{table}
 \begin{tabular}{ lccc }
 \hline
Target& Number of & F1 Score& F1 Score\\
Dataset& Clusters &No Active Learning&All 1,280 points\\
 \hline
\multirow{5}{*}{AWS} &1 	&  0.3096	& \textbf{0.6161}\\ 
 				&2  	&  0.3010	& 0.3474	\\ 
				&3 	&  0.3159 & 0.2965	\\ 
				&5  	&  0.3053	& 0.1956	\\ 
				&10	&  \textbf{0.3162} & 0.3199\\ 
 \hline
\multirow{5}{*}{Huawei} &1 &  0.3664 & \textbf{0.8626}\\ 
 				&2  	&  0.3595	& 0.4276	\\ 
				&3 	& \textbf{0.3676} & 0.4537	\\ 
				&5  	&  0.1777	& 0.1540	\\ 
				&10	&  0.2669  & 0.2584	\\ 
 \hline
\multirow{5}{*}{IOPS} &1 	&  0.3259	& \textbf{0.4946}	\\ 
 				&2  	&  \textbf{0.3514} & 0.3629\\ 
				&3 	&  0.3367	& 0.2691\\ 
				&5  	&  0.3434	& 0.3847	\\ 
				&10	&   0.3083	& 0.3123	\\ 
\hline
\multirow{5}{*}{Twitter} &1 	& 0.1277 	& \textbf{0.2129}\\ 
 				&2  	&  0.1611	& 0.1275	\\ 
				&3 	& \textbf{0.1775} & 0.1747	\\ 
				&5  	&  0.1774	& 0.1760	\\ 
				&10	&   0.1746 & 0.1479	\\ 
\hline
\multirow{5}{*}{Yahoo Artificial} &1 	& 0.7427 	& \textbf{0.9516}	\\ 
 				&2  	&  0.4974	& 0.8082	\\ 
				&3 	& \textbf{0.8603} & 0.8124	\\ 
				&5  	&  0.8570	& 0.8415	\\ 
				&10	&   0.8440 & 0.7681	\\ 
\hline
\multirow{5}{*}{Yahoo Real} &1 	&  0.5834	& \textbf{0.7918}	\\ 
 				&2  	&  0.5871	& 0.6240	\\ 
				&3 	&  \textbf{0.6354} & 0.6601	\\ 
				&5  	&  0.6260	& 0.6738	\\ 
				&10	&  0.5713  & 0.5698\\ 
\hline
\end{tabular}  
\caption{For each transfer learning dataset combination the mean (across 5 folds) F1 scores for 1, 2, 3, 5 and 10 clusters (a) before the active learning process is run, and (b) after active learning has been run to label 1,280 points. For each dataset, the best score before and after active learning is highlighted in bold font.}
\label{tab:exp1f1bycluster}
\end{table}

\section{Experiment 2: Assessing the rate of improvement in model performance using active learning}
\label{sec:exp2}

Experiment  1 (Section~\ref{sec:exp1}) focused on exploring the interaction between the number of clusters used during transfer learning and active learning. In this second experiment, we shift focus to the rate of improvement in model performance as active learning is applied to select new samples for inclusion in the training data. Given the results from experiment 1 indicated that the best performance with active learning is achieved when a single cluster is used for transfer learning (note this is equivalent to having no clustering in the data processing pipeline) for this experiment we restrict our analysis to the single cluster results. 

\begin{figure}
	\includegraphics[width=\textwidth]{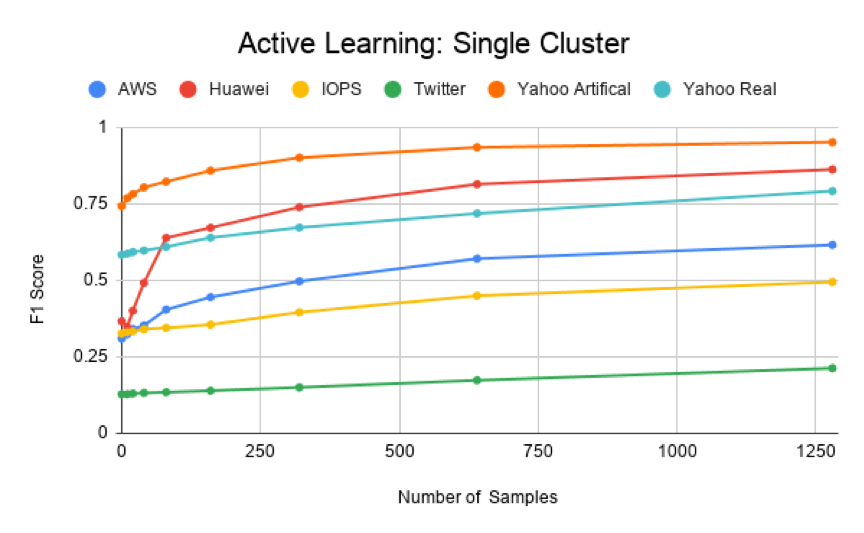}
	\caption{Plots of the improvement in F1 for each dataset as more points are added using active learning}
	\label{fig:exp2f1trajectories}
\end{figure}

Figure~\ref{fig:exp2f1trajectories} shows a plot of the improvement in model performance for each of the 6 datasets as the number of points added to the training set using active learning increases from 0 through to 1280. Table~\ref{tab:exp2performancetrajectories1} and Table~\ref{tab:exp2performancetrajectories2} provide more detailed information on this process and present the precision, recall and F1 of the system on the test set for each (target) dataset as the number of points from the target domain training split are added to the model training set via active learning.

From Figure~\ref{fig:exp2f1trajectories} it is apparent that the performance of the system varies significantly across the datasets. For example, the worst performance is on the Twitter dataset. The bad performance on Twitter is likely because Twitter is the smallest dataset and it is also the least similar to the other dataset from a domain perspective (the other datasets contain cloud architecture monitoring data while the Twitter data record the volume of tweets). 

Focusing on the impact of active learning on model performance as more points are added to the training data, two observations can be made: (a) in all cases model performance improves as the active learning process progresses, and (b) excluding the Huawei dataset the improvement in model performance is relatively linear and quite small across the number of points added. It is difficult to directly compare these results to the results reported by \cite{zhang2019cross} for two reasons: first, they do not report the performance of their base model on each dataset before active learning; and, second, when they apply active learning they sample points directly from the test set (resulting in the test set becoming smaller and less difficult as active learning progresses. We can however attempt a general comparison to put our results in context. On the Yahoo-real dataset, the results reported in \cite{zhang2019cross} are that a Random Forest model trained on non-Yahoo-real data and a sample of approximately 60 data points randomly sampled from the Yahoo-real domain achieves an F1 score of 0.3348, they also report that using their full system model that after 920 points are sampled from Yahoo-real (using active learning) they achieved an F1 of 0.5697. For comparison, we will treat the random forest model from \cite{zhang2019cross} which was given access to some target training data as equivalent to our base model before active learning (i.e., our model that did not have any Yahoo-real data in its training). Using this assumption of base model equivalence Table~\ref{tab:exp2yahoozhang} lists the F1 scores for the baseline models on transferring to Yahoo-real, the system performance after active learning has run, the number of points sampled through the active learning process and the average improvement in F1 per active learning point added. Table~\ref{tab:exp2awszhang} presents a similar analysis and comparison for the AWS dataset. 

The analysis presented in Table~\ref{tab:exp2yahoozhang} and Table~\ref{tab:exp2awszhang} reveals that the rate of performance in system performance using active learning is relatively stable across both these datasets. For both datasets, the rate of improvement in performance in F1 is higher in the results reported by \cite{zhang2019cross}. However, this is likely because they sample points directly from the test set during active learning resulting in the test set becoming easier as active learning progresses because the points the model finds most difficult are removed for training purposes through the active learning process. The analysis also shows that measured in terms of improvement in F1 per data point the impact of active learning is relatively small (the most significant digit being in the range of $10^{-4}$). Although it should be added that the Huawei dataset is an outlier in this regard, active learning significantly positively impacted performance over the first 100 samples.

\begin{table}
 \begin{tabular}{ lcccc }
 \hline
Target& Active Learning & ~ & ~ & ~\\
Dataset& Points & Precision &Recall & F1\\
 \hline
\multirow{9}{*}{AWS} &0 	&  0.5782	& 0.2114 & 0.3096\\ 
& 10	& 0.6122	& 0.2203	& 0.3240\\
& 20	& 0.6635	& 0.2296	& 0.3411\\
& 40	& 0.7458	& 0.2308	& 0.3525\\
& 80	& 0.8144	& 0.2690	& 0.4044\\
& 160& 	0.8575	& 0.3006	& 0.4451\\
& 320& 	0.8985	& 0.3435	& 0.4970\\
& 640& 	0.9373	& 0.4104	& 0.5709\\
& 1280& 	0.9557	& 0.4546	& 0.6161\\
 \hline
\multirow{9}{*}{Huawei} & 0		& 0.5119	& 	0.2853	& 	0.3664\\
&10	& 	0.5188	& 	0.2614	& 	0.3476\\
&20	& 	0.5821	& 	0.3054	& 	0.4006\\
&40	& 	0.6675	& 	0.3892	& 	0.4917\\
&80	& 	0.7710	& 	0.5455	& 	0.6390\\
&160	& 	0.8182	& 	0.5702	& 	0.6720\\
&320	& 	0.9314	& 	0.6127	& 	0.7391\\
&640	& 	0.9808	& 	0.6961	& 	0.8143\\
&1280		& 0.9892	& 	0.7648	& 	0.8626\\
 \hline
\multirow{9}{*}{IOPS} &0 & 	0.5875 & 	0.2255 & 	0.3259\\
&10 & 	0.5920 & 	0.2294 & 	0.3307\\
&20 & 	0.6053 & 	0.2298 & 	0.3331\\
&40 & 	0.6403 & 	0.2317 & 	0.3402\\
&80 & 	0.6426 & 	0.2355 & 	0.3447\\
&160 & 	0.6577 & 	0.2434 & 	0.3553\\
&320 & 	0.6972 & 	0.2763 & 	0.3957\\
&640 & 	0.7830 & 	0.3156 & 	0.4498\\
&1280 & 	0.9135 & 	0.3391 & 	0.4946\\
\hline
\end{tabular}  
\caption{The precision, recall and F1 of the system on the test set for the AWS, Huawei and IOPS (target) datasets as the number of points from the target domain training split are added to the model training set via active learning}
\label{tab:exp2performancetrajectories1}
\end{table}

\begin{table}
 \begin{tabular}{ lcccc }
 \hline
Target& Active Learning & ~ & ~ & ~\\
Dataset& Points & Precision &Recall & F1\\
 \hline
\multirow{9}{*}{Twitter} &0	&0.4890 &	0.0734 &	0.1277\\
&10 &	0.4881 &	0.0735 &	0.1278\\
&20 &	0.4977 &	0.0748 &	0.1300\\
&40 &	0.5088 &	0.0759 &	0.1321\\
&80 &	0.5422 &	0.0767 &	0.1344\\
&160	 &0.6415 &	0.0782 &	0.1394\\
&320	 &0.7184 &	0.0838 &	0.1501\\
&640	 &0.8305 &	0.0968 &	0.1733\\
&1280 &	0.8892 &	0.1209 &	0.2129\\
\hline
\multirow{9}{*}{Yahoo Artificial} &0 &		0.9987 &		0.5912 &		0.7427\\
&10 &		0.9990 &		0.6251 &		0.7690\\
&20 &		0.9985 &		0.6420 &		0.7815\\
&40 &		0.9988 &		0.6723 &		0.8036\\
&80 &		0.9991 &		0.6990 &		0.8225\\
&160	 &	0.9988 &		0.7531 &		0.8587\\
&320	 &	0.9991 &		0.8204 &		0.9010\\
&640	 &	0.9991 &		0.8783 &		0.9348\\
&1280 &		0.9991 &		0.9084 &	0.9516\\
\hline
\multirow{9}{*}{Yahoo Real} &0	&0.8446&	0.4456&	0.5834\\
&10&	0.8416&	0.4509&	0.5872\\
&20&	0.8431&	0.4568&	0.5925\\
&40&	0.8551&	0.4593&	0.5976\\
&80&	0.8682&	0.4697&	0.6096\\
&160	&0.8816&	0.5019&	0.6396\\
&320	&0.9100&	0.5333&	0.6725\\
&640	&0.9421&	0.5809&	0.7187\\
&1280&	0.9670&	0.6704&	0.7918\\
\hline
\end{tabular}  
\caption{The precision, recall and F1 of the system on the test set for the Twitter, Yahoo Artificial and Yahoo Real datasets as the number of points from the target domain training split are added to the model training set via active learning}
\label{tab:exp2performancetrajectories2}
\end{table}

\begin{table}
 \begin{tabular}{ lccccc}
 \hline
~ & F1 Before & F1 After & ~ & \#Points & Per Point \\
Yahoo-real& Active & Active & Difference & Added & Increase\\
~ & Learning & Learning & In F1 & via AL & in F1\\	
\hline
Model reported in \cite{zhang2019cross} & 	0.3348 & 	0.5697 & 	0.2349 & 	920 & 0.00026\\
Model from this Experiment & 0.5834 & 0.7918& 0.2084 & 1280 &	0.00016\\
\hline
\end{tabular}  
\caption{An analysis of the improvement in F1 performance on the Yahoo-real dataset using active learning (AL)}
\label{tab:exp2yahoozhang}
\end{table}

\begin{table}
 \begin{tabular}{ lccccc}
 \hline
~ & F1 Before & F1 After & ~ & \#Points & Per Point \\
AWS & Active & Active & Difference & Added & Increase\\
~ & Learning & Learning & In F1 & via AL & in F1\\	
\hline
Model reported in \cite{zhang2019cross} & 0.7974 &	0.8637 &	0.0663 &	254	& 0.00026\\
Model from this Experiment & 0.3096 &	0.6161 &	0.3065 &	1280	& 0.00013\\
\hline
\end{tabular}  
\caption{An analysis of the improvement in F1 performance on the AWS dataset using active learning (AL)}
\label{tab:exp2awszhang}
\end{table}

\section{Experiment 3: Assessing Ceiling Performance for Active Learning}
\label{sec:exp3}

Experiment 2 (Section~\ref{sec:exp3}) assessed the rate of improvement of system performance in the relatively early stages of active learning (i.e. up to 1280 points being sampled). In this third experiment, we assess how model performance varies as the number of points sampled is increased substantially, and also compare the optimal performance of a transfer and active learning approach compared to in-domain training. 

As the datasets are of varying sizes the values we report active learning performance for are defined as percentages of the set of data that points are being selected from (target training data). Note that the target training dataset's size is proportional to the target test set's size. Also, IOPS is a very large dataset and so for computational reasons, it was not feasible to report results for IOPS up to 100\% of the target training dataset. For this reason, IOPS was excluded from this experiment. 

\begin{figure}
	\includegraphics{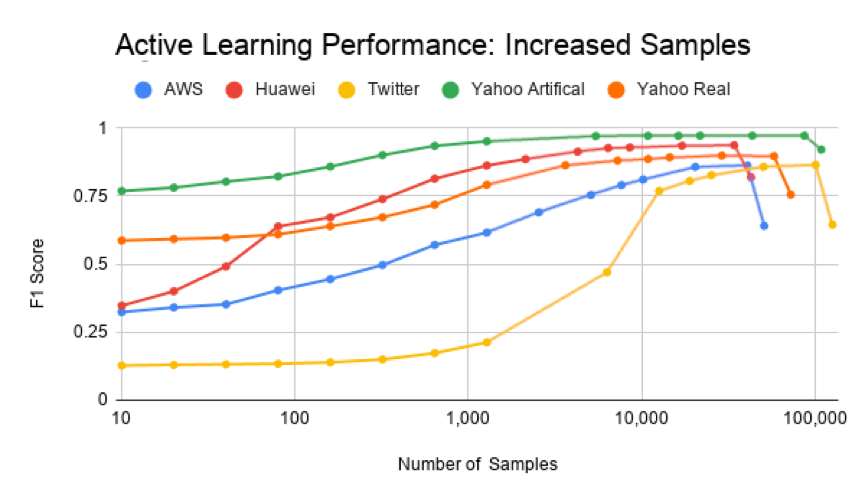}
	\caption{Plots of the improvement in F1 for each dataset as more points are added using active learning, in this instance up to 100\% of the target domain training set.}
	\label{fig:exp3}
\end{figure}

Figure~\ref{fig:exp3} plots the improvement in model performance across the 5 datasets from a baseline of transfer learning as active learning progresses up to 100\% of the target training set split. Table~\ref{tab:exp3} lists the overall dataset system performance as the number of points sampled from the target training dataset through active learning increases. To provide a comparator between a transfer learning and active learning combination versus and within domain training approach the rightmost column of the table reports the mean precision, recall and F1 scores across a 5-fold cross-validation for a system trained solely on the dataset.

\begin{table}
 \begin{tabular}{ ll|cccc|c}
 \hline
~ & ~ & \multicolumn{4}{c|}{Non-X $\rightarrow$ X Transfer \&} & Within\\
~ & ~ & \multicolumn{4}{c|}{Active Learning Performance} & Domain\\
~ & ~ & \@ 20\% & \@ 40\% & \@ 80\% & \@ 100\% & Performance\\	
\hline
\multirow{3}{*}{AWS} & Precision & 0.9876	& 0.9895 &	0.9877 &	0.9512 & 	0.9938\\
~ & Recall &0.6885 &	 0.7570 &	0.7653 &	0.4831 & 	0.8922\\
~ & F1 &0.8113 & 	0.8578 & 0.8624 & 0.6407 & 0.9403\\
\hline
\multirow{3}{*}{Huawei} & Precision &  0.9927	&0.9923	&0.9940	&0.9450	&0.9964\\
~ & Recall & 0.8733	&0.8847	&0.8869	&0.7231	&0.9575\\
~ & F1 & 0.9292	&0.9354	&0.9374	&0.8193	&0.9766\\
\hline
\multirow{3}{*}{Twitter} & Precision &  0.9619 &	0.9659 & 	0.9681 &	0.9219 &	0.9773\\
~ & Recall & 0.7262 &	0.7715 &	0.7819 & 	0.4961 &	0.8987\\
~ & F1 & 0.8276 &	0.8578 &	0.8651 &	0.6451 &	0.9364\\
\hline
\multirow{3}{*}{Yahoo Artificial} & Precision & 0.9993 & 0.9991&	0.9992&	0.9992&	0.9991\\
~ & Recall & 0.9471&	0.9471&	0.9471&	0.8545 &	0.9864\\
~ & F1 & 0.9725&	0.9724&	0.9724&	0.9212 &	0.9927\\
\hline
\multirow{3}{*}{Yahoo Real} & Precision & 0.9825&	0.9823&	0.9816&	0.9383&	0.9820\\
~ & Recall & 0.8179&	0.8290&	0.8260&	0.6327&	0.9236\\
~ & F1 & 0.8927&	0.8992&	0.8971&	0.7558&	0.9519\\
\hline
\end{tabular}  
\caption{Per dataset system performance as the number of points sampled from the target training dataset are sampled by active learning and in the rightmost column the mean precision, recall and F1 scores across a 5-fold cross-validation for a system trained solely on the target dataset (no transfer learning or active learning)}
\label{tab:exp3}
\end{table}

The plots in Figure~\ref{fig:exp3} make it apparent that system performance continues to improve in all datasets using active learning up to a significant number ($>10,000$) samples being selected by active learning. However, a consistent phenomenon across all datasets is system performance does not saturate as active learning progresses, instead, it eventually begins to deteriorate. A potential explanation for this is that active learning is doing a good job of selecting useful data points to add to the training data and that as we approach 100\% of the pool of samples that active learning is sampling from, the active learning selection process is overridden as all remaining points are added, and this results in data points that are detrimental to learning being added (for example, these might be the points in the target domain training sample that are most similar to the source domain datasets and least representative of the target domain and that by adding these points to the training data it shifts the distribution learned by the model back towards the source domain). 

Examining the results listed in Table~\ref{tab:exp3} an interesting observation is that transfer learning combined with active learning can outperform systems that are trained purely on in-domain data. For all datasets, the best performance is achieved with a combination of transfer learning and active learning as compared with purely domain training.  Note that the metrics reported for the in-domain training are the average across 5 folds, and so they are equivalent to using 80\% of the target domain dataset for training. Given this, it is interesting that for some datasets the transfer learning and active learning combination outperforms the in-domain training using less in-domain training data. 

\section{Conclusions}

To conclude our results indicate that although transfer learning in isolation can benefit from using multiple clusters to identify sub-domains within multi-source domains, when used in combination with active learning best performance is achieved when clustering is removed from the process. Furthermore, the per sample improvement in model performance using active learning is relatively stable across datasets but is also relatively small. Although it should be added that the Huawei dataset is an outlier in this regard, active learning significantly positively impacted performance over the first 100 samples. Finally, transfer learning in combination with active learning can outperform purely within target-domain training and can, in some cases, achieve this superior performance using less target domain data. Admittedly, however, achieving this performance still requires a significant amount of target data to be labelled.

\section*{Acknowledgements}
This research was supported by the ADAPT Research Centre. The ADAPT Centre for Digital Content Technology is funded under the Research Ireland's Research Centres Programme (Grant 13/RC/2106\_II) and is co-funded under the European Regional Development Funds.

\bibliography{TSTL-Bibliography}
\bibliographystyle{plain}

\end{document}